\documentclass[10pt,twocolumn,letterpaper]{article}

\usepackage[pagenumbers]{cvpr} 

\usepackage[dvipsnames]{xcolor}

\newcommand{\nothing}[1]{}

\definecolor{cvprblue}{rgb}{0.21,0.49,0.74}
\usepackage[pagebackref,breaklinks,colorlinks,citecolor=cvprblue]{hyperref}
\usepackage{multirow}
\usepackage{cuted}
\usepackage{marvosym}

\def\paperID{1313} 
\def\confName{CVPR}
\def\confYear{2024}

\title{HAVE-FUN: Human Avatar Reconstruction from \\ Few-Shot Unconstrained Images}

\author{Xihe Yang\textsuperscript{1,2,\textdagger*} \hspace{0.15in}
Xingyu Chen\textsuperscript{1,\Letter}\thanks{Equal contribution; \Letter\,Corresponding author.} \hspace{0.15in}
Daiheng Gao\textsuperscript{4} \hspace{0.15in} 
Shaohui Wang\textsuperscript{1,5,}\thanks{This work was done when Xihe~Yang and Shaohui~Wang were interns at Xiaobing.AI, led by Xingyu Chen.} \\
Xiaoguang Han\textsuperscript{2,3} \hspace{0.15in}
Baoyuan Wang\textsuperscript{1,\Letter} \\
\textsuperscript{1}Xiaobing.AI \hspace{0.15in} 
\textsuperscript{2} SSE, CUHKSZ \hspace{0.15in} 
\textsuperscript{3} FNii, CUHKSZ \hspace{0.15in} 
\textsuperscript{4}Freelancer \hspace{0.15in} 
\textsuperscript{5}Tsinghua University  \\
\url{https://seanchenxy.github.io/HaveFunWeb/}
}

\begin{document}
\maketitle
\begin{strip}
    \centering
    \vspace{-50pt}
    \includegraphics[width=\linewidth]{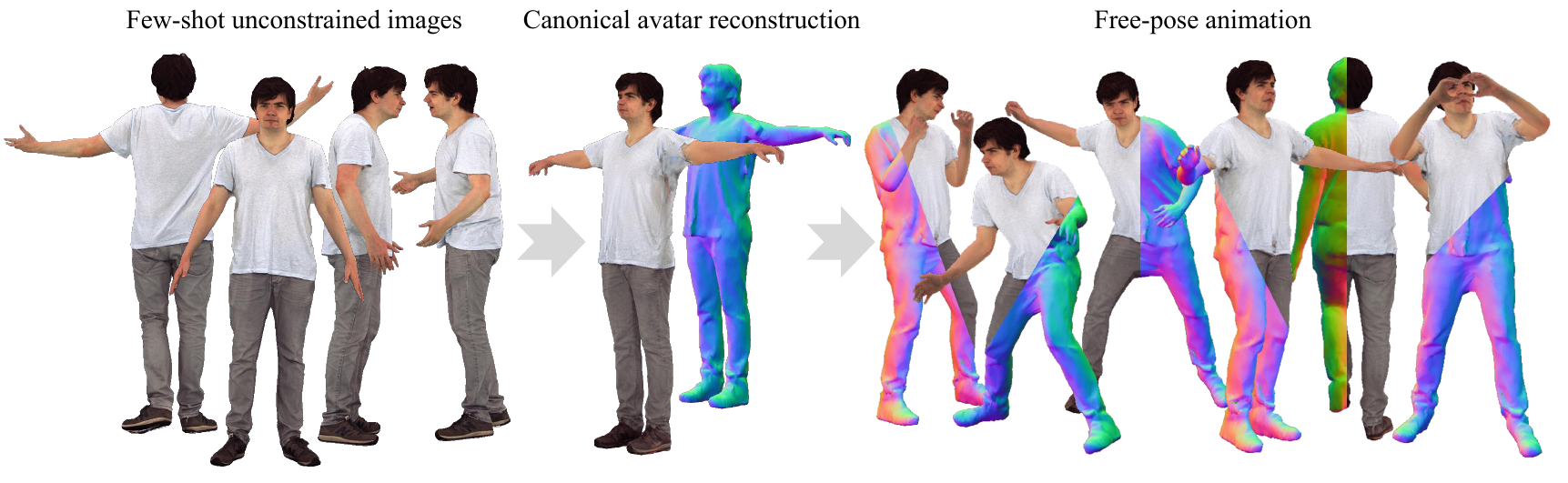}\\
    \captionsetup{type=figure,font=small}
    \caption{Given a few images with various viewpoints and articulated poses, our approach can reconstruct an animatable human avatar.
    }
    \label{fig:teaser}
\end{strip}
\begin{abstract}
As for human avatar reconstruction, contemporary techniques commonly necessitate the acquisition of costly data and struggle to achieve satisfactory results from a small number of casual images. In this paper, we investigate this task from a few-shot unconstrained photo album. The reconstruction of human avatars from such data sources is challenging because of limited data amount and dynamic articulated poses.  For handling dynamic data, we integrate a skinning mechanism with deep marching tetrahedra (DMTet) to form a drivable tetrahedral representation, which drives arbitrary mesh topologies generated by the DMTet for the adaptation of unconstrained images. To effectively mine instructive information from few-shot data, we devise a two-phase optimization method with few-shot reference and few-shot guidance. The former focuses on aligning avatar identity with reference images, while the latter aims to generate plausible appearances for unseen regions. Overall, our framework, called HaveFun, can undertake avatar reconstruction, rendering, and animation. Extensive experiments on our developed benchmarks demonstrate that HaveFun exhibits substantially superior performance in reconstructing the human body and hand. 

\end{abstract}    
\section{Introduction}
\label{sec:intro}
Human avatar reconstruction has experienced rapid development in recent years and shown great potential for applications of AR/VR, metaverse, etc \cite{bib:handAR,bib:mobrecon}. 
One of the challenges in this field is the data acquisition. Previous works typically require an expensive setup for multi-view RGB video \cite{bib:NeuralBody}, textured scan video \cite{bib:XAvatar}, or multi-view static images \cite{bib:PIFu}. 
Recently, there is a tendency to utilize easily accessible data sources for this task, such as monocular RGB video \cite{bib:HumanNeRF} or image set \cite{bib:PersonNeRF}.

This leads to a pivotal question: ``Is it possible to leverage a cheaper data source to reconstruct human avatars?'' Intuitively, the cheaper data should be characterized by its limited quantity and the unconstrained nature of human behaviors it captures. As this paper will demonstrate, the answer leans toward the affirmative, and we refer to such a data source as few-shot unconstrained images. These data can be obtained from either a personal photo album or key frames of a video capture. Thereby, we are inspired to explore the human avatar reconstruction task with the few-shot unconstrained images.



From a technical perspective, we carefully design a 3D representation for the difficulties caused by the aforementioned data setting. When it comes to modeling dynamic humans, a popular idea is the use of a dynamic neural radiance field (NeRF) \cite{bib:NeRF}. Nevertheless, despite various dynamic designs being reported \cite{bib:DNeRF,bib:HumanNeRF,bib:SNARF,bib:SANeRF}, accurately driving a volume space using limited data remains challenging.
Recently, deep marching tetrahedra (DMTet) \cite{bib:DMTet} proposes a hybrid method to produce a triangle mesh through a differentiable process. Due to the ease of mesh deformation, we are motivated to model dynamic humans with the DMTet representation. Specifically, we integrate skinning weights and blendshapes defined by SMPLX \cite{bib:SMPLX} with the static DMTet to create a drivable tetrahedral representation for adapting the unconstrained data. 
Furthermore, conducive to the few-shot task, the driveable representation allows the use of canonical SMPLX shape as the prior to make the tetrahedral grid well-initialized. In this manner, we extend the traditional static-scene DMTet to model the articulated human body under a dynamic condition.

In terms of learning from a limited amount of data, we employ the score distillation sampling (SDS) \cite{bib:DreamFusion,bib:SJC} technique to generate plausible textures for unseen regions. Different from existing SDS-based image-to-3D tasks that rely on textual descriptions or captions \cite{bib:RealFusion,bib:MakeIt3D}, we directly utilize the image as the prompt \cite{bib:Zero123} so that faithful visual feature can be preserved. We refer to the SDS-based optimization as a few-shot guidance. Additionally, we incorporate traditional reconstruction optimization, namely few-shot reference, in our pipeline.

Our overall framework is called HaveFun, \textbf{H}uman \textbf{AV}atar r\textbf{E}construction from \textbf{F}ew-shot \textbf{UN}constrained images. Note that the human body and hand have distinct properties. The body is characterized by intricate geometry (\eg, hair/cloth) and remarkable facial features, while the hand exhibits smooth bare geometry and subtle palm wrinkles. To evaluate our framework for the human body and hand, we develop benchmarks for them with the assistance of XHumans \cite{bib:XAvatar} and DART \cite{bib:DART}.  
Remarkably, HaveFun effectively addresses both scenarios of the human body and hand. As a result, our approach can reconstruct human avatars with few-shot (as few as 2) dynamic images and achieve realistic rendering quality. Besides, we can perform avatar animation in various unseen human poses. 

Our main contributions are summarized as follows:
\begin{itemize}[leftmargin=*,topsep=3pt]
    \setlength\itemsep{0em}
    \item 
    We propose a novel framework, termed HaveFun, to solve the challenging problem of human avatar reconstruction from few-shot unconstrained images.
    \item We explore a drivable tetrahedral representation for articulated human motion and an SDS loss for non-static human reconstruction. 
    \item We develop benchmarks for the few-shot dynamic body/hand reconstruction task. Extensive evaluations indicate our method outperforms previous one-shot \cite{bib:TeCH} or video-based \cite{bib:SelfRecon} approaches by a large margin. 
\end{itemize}
We believe our endeavors would 
enhance the practical significance of this research area, paving a new way for human avatar reconstruction and real-world applications.

\section{Related Work}
\label{sec:related_work}

\paragraph{Avatar reconstruction of the human body.}
Prominent techniques for human body reconstruction necessitate expensive data acquisition, \eg, multi-view RGB video \cite{bib:NeuralBody,bib:SANeRF,bib:SLRF,bib:AvatarRex,bib:PoseVocab,bib:HumanRF,bib:HandNeRF,bib:RelightableHands,bib:LISA,bib:DoubleField,bib:HumanNeRFSP,bib:AnimNeRF,bib:ARCH,bib:TAVA}, monocular RGB video \cite{bib:HumanNeRF,bib:IMAvatar,bib:HandAvatar,bib:NHA,bib:Vid2avatar,bib:MonoHuman,bib:HARP,bib:Te,bib:SelfRecon,bib:NeuMan}, textured scan video \cite{bib:XAvatar,bib:LocalEdit}, or image set \cite{bib:PersonNeRF}. Also, many of them utilize NeRF or its variations as the 3D representation and craft a motion field to connect the gap between body articulation and the canonical NeRF space.
For example, HumanNeRF \cite{bib:HumanNeRF} created a personalized avatar by training an inverse skinning field using monocular video data. LISA \cite{bib:LISA} utilized multi-view video data to learn hand appearance and incorporated a multi-layer perceptron (MLP) to forecast skinning weights for hand animation. PersonNeRF \cite{bib:PersonNeRF} gathered hundreds of images of a human individual and generated an avatar with disentangled attributes. 
Differing from the majority of prior studies, our focus lies in addressing the few-shot body reconstruction challenge. Moreover, we explore articulation-friendly tetrahedral grid as the 3D representation.

\vspace{-0.4cm}
\paragraph{Few-shot human avatar creation.}
Many research efforts rely on pre-trained generative models \cite{bib:EG3D,bib:AG3D,bib:HumanGen,bib:Next3D,bib:Get3DHuman,bib:Mimic3D,bib:GRAM} and accomplish one-shot reconstructions through GAN inversion techniques \cite{bib:PTI,bib:GRAMInv}.
This line of research typically requires only a single image to recover a latent representation that aligns with the ground truth.
Nevertheless, these pipelines often struggle to accurately represent the data that falls outside the GAN distribution \cite{bib:img2stylegan,bib:HGANinv}.
Instead of GAN inversion, another approach involves directly extracting image features and predicting a pixel-aligned implicit field to represent the human \cite{bib:S3F,bib:ICON,bib:ECON,bib:PIFu,bib:PIFuHD}. 
As a groundbreaking work, PIFu \cite{bib:PIFu} employed MLPs to model both the occupancy value and color of the human body from one or several images.
Despite the few-shot setting in the inference phase, the training of pixel-aligned methods still demands a large-size image set.
Further, they treat the human as a static scene, neglecting the dynamic nature of the human body. By contrast, this paper can handle few-shot dynamic human images without the need for additional auxiliary data collection.

\vspace{-0.4cm}
\paragraph{Sparse-view 3D reconstruction.}
When it comes to few-shot reconstruction, traditional methods typically optimize a NeRF using sparse-view data with the aid of geometry regularization, semantic consistency, depth supervision, \etc. \cite{bib:RegNeRF,bib:DSNeRF,bib:PSNeRF,bib:SparseNeuS,bib:DietNeRF,bib:InfoNeRF}. For example, 
RegNeRF \cite{bib:RegNeRF} introduced a patch regularizer to mitigate geometry artifacts and employed a log-likelihood model to ensure multi-view appearance consistency. 
In contrast to this line, we utilize a pre-trained diffusion model to regularize unseen textures from novel viewpoints. Moreover, in contrast to static scenes typically handled by related works, our approach enables dynamic body reconstruction using sparse-view data.

\vspace{-0.4cm}
\paragraph{Avatar creation with text-based priors.}
Recently, text-to-3D task has gained popularity, thanks to the pre-trained language-vision models \cite{bib:DDPM,bib:CLIP}. Many reports have achieved remarkable performance using the diffusion model \cite{bib:DreamFusion,bib:Magic3D,bib:MakeIt3D,bib:RealFusion,bib:SJC}. As a pioneering effort, DreamFusion \cite{bib:DreamFusion} proposed SDS loss to optimize a NeRF with text prompts. Furthermore, text-guided strategies have been incorporated into the field of human avatar creation
\cite{bib:AvatarCraft,bib:DreamHuman,bib:DreamAvatar,bib:TeCH,bib:TADA,bib:AvatarVerse,bib:DreamWaltz,bib:ELICIT,bib:AvatarCLIP}. These approaches allow for the generation of a realistic 3D human appearance that aligns with text semantics. While the text-based paradigm is effective in creating famous characters, it is more challenging when it comes to reconstructing actual human individuals. 
For example, TeCH \cite{bib:TeCH} accomplished one-shot human avatar reconstruction 
with 5 stages of VQA \cite{bib:BLIP} caption, DreamBooth \cite{bib:DreamBooth} fine-tuning, geometry optimization, geometry post-processing, and texture optimization. In contrast, our method can be trained in an end-to-end manner.
Hence, we believe our approach is inherently more elegant, precise, and accurate than text-based methods for the reconstruction task since we directly employ image features as guidance without the potential ambiguity caused by image captions.

\section{Approach}

\begin{figure*}[t]
\begin{center}
\includegraphics[width=\linewidth]{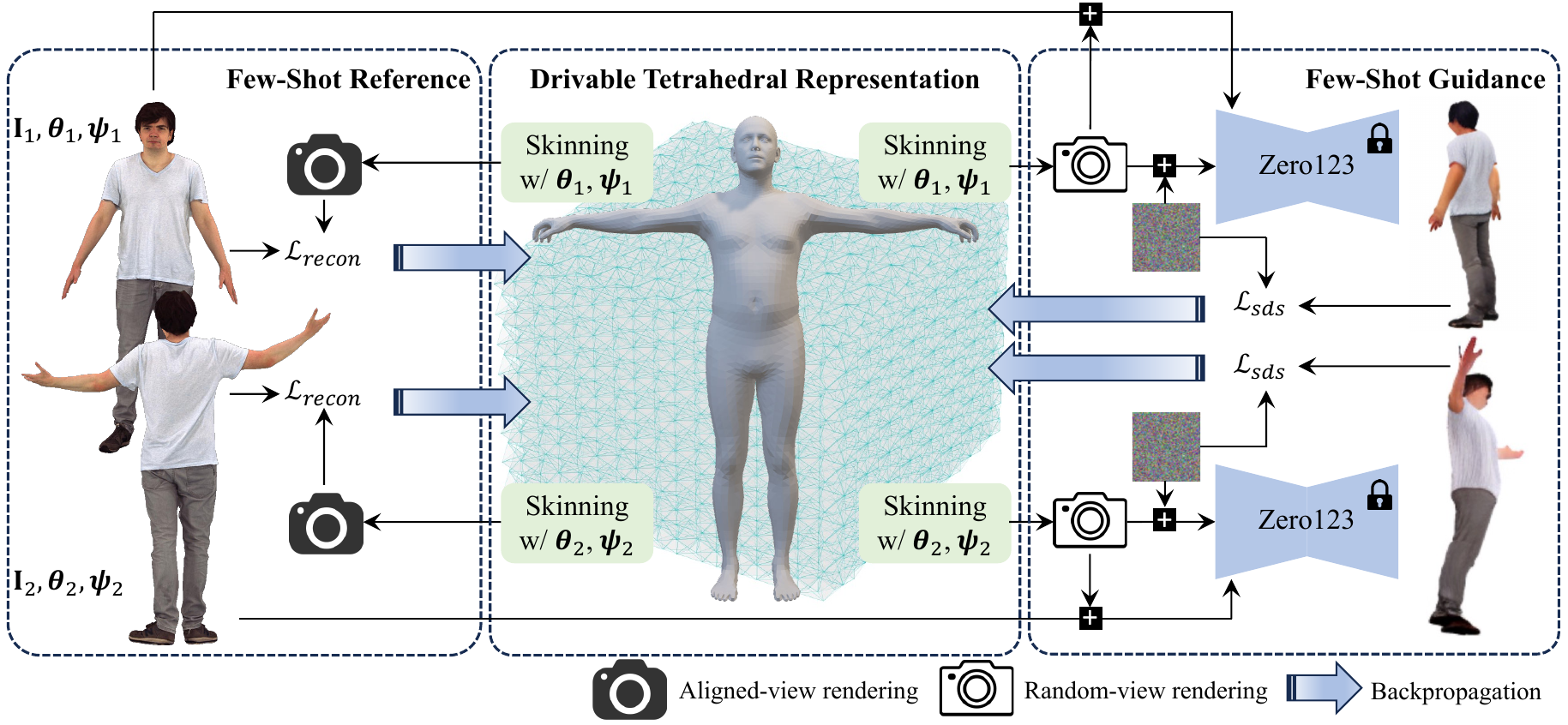}
\caption{
Overview of HaveFun framework. Based on the DMTet, we design a driveable tetrahedral representation with the skinning mechanism. In terms of optimization, we employ loss functions based on reference-data reconstruction and SDS guidance to create human avatars from few-shot unconstrained images.
}
\label{fig:arch}
\end{center}
\vspace{-0.3cm}
\end{figure*}

Given a personalized unconstrained photo album $\mathcal I=\{\mathbf I_i\}_{i=1}^N (N\leq 8)$, this paper aims to reconstruct a 3D representation $\mathcal G$ for free-viewpoint rendering and free-pose animation. $\mathcal G$ takes explicit viewpoint $\mathbf R\in\mathbb R^{d_{\mathbf R}}$, human articulated poses $\boldsymbol\theta\in\mathbb R^{d_{\boldsymbol\theta}}$, and expression coefficients $\boldsymbol\psi\in\mathbb R^{d_{\boldsymbol\psi}}$ as the input and generates a 2D image $\hat{\mathbf I}$:
\begin{equation}
\label{eq:g}
\mathcal G: (\mathbf R,\boldsymbol\theta,\boldsymbol\psi)\in\mathbb R^{d_{\mathbf R}}\times\mathbb R^{d_{\boldsymbol\theta}}\times \mathbb R^{d_{\boldsymbol\psi}} \rightarrow \hat{\mathbf I}\in\mathbb R^{H\times W\times 3},
\end{equation}
where $H=W=512$ denote the image height and width. The 3D representation $\mathcal G=\{\mathcal M,\mathcal C,\mathbf W,\mathbf E\}$ includes a triangular mesh $\mathcal M$, a texture field $\mathcal C$, skinning weights $\mathbf W$, and expression blendshapes $\mathbf E$. As shown in Fig.~\ref{fig:arch}, we build a drivable tetrahedral representation as the core of $\mathcal G$ to produce $\{\mathcal M,\mathcal C,\mathbf W, \mathbf E\}$ (Sec.~\ref{sec:tet}). Additionally, we design two phases of few-shot reference and few-shot guidance to train our framework (Sec.~\ref{sec:opt}). We describe each part in detail below.

\subsection{Preliminaries}
\label{sec:pre}

\paragraph{Deep marching tetrahedra.}
DMTet is a hybrid representation for 3D geometry, denoted as $(\mathbf V^t, \mathbf T)$, where $\mathbf V^t,\mathbf T$ are vertices and the tetrahedral indices, respectively. Each tetrahedron $\mathbf t \in \mathbf T$ is represented by four vertices $\{\mathbf v^t_a,\mathbf v^t_b,\mathbf v^t_c,\mathbf v^t_d\}$.
Each vertex $\mathbf v^t=(x,y,z) \in \mathbf V^t$ has a 3D position vector and a signed distance value $s$. If two vertices in a tetrahedron have different signs of $s$ (\eg, $\mathbf v^t_a$ with $s_a<0$ and $\mathbf v^t_b$ with $s_b>0$), they can determine a vertex $\mathbf v^m$ in triangular mesh $\mathcal M$:
\begin{equation}
\label{eq:dmtet}
\mathbf{v}^m = \frac{(\mathbf v^t_a+\delta\mathbf v^t_a)\cdot s^t_b-(\mathbf v^t_b+\delta\mathbf v^t_b)\cdot s_a}{s_b-s_a},
\end{equation}
where $\delta\mathbf{v}$ represents the estimated vertex displacement. As a result, the triangular mesh can be generated using the differentiable volume subdivision method. However, the DMTet cannot represent an articulated dynamic object.

\vspace{-0.4cm}
\paragraph{Viewpoint-conditioned diffusion model.}
Zero123 \cite{bib:Zero123} and follow-up works \cite{bib:consistent123,bib:Zero123++,bib:syncdreamer,bib:one2345} introduce a denoising diffusion model that leverage posed CLIP embedding, involving both the visual CLIP feature \cite{bib:CLIP} and the camera viewpoint $\delta\mathbf R$, as the conditioning elements.
Consistent with the traditional diffusion model \cite{bib:DDPM}, Zero123 has a forward sampling process:
\begin{equation}
\label{eq:forward}
\mathbf z_t=\sqrt{\bar\alpha_t}\mathbf I+\sqrt{(1-\bar\alpha_t)}\epsilon, \epsilon\sim \mathcal{N}(0,1),
\end{equation}
where $\bar\alpha_t$ is a hyperparameter and $\mathbf z_t$ is noising image at the $t$-th step.
Subsequently, the noise prediction model $\hat\epsilon$ is optimized to estimate the added noise $\epsilon$:
\begin{equation}
\label{eq:0123loss}
\min \mathbb{E}_{t, \epsilon}\left\|\epsilon-\hat\epsilon\left(\mathbf z_t, t, \mathrm{CLIP}(\mathbf I), \delta\mathbf R)\right)\right\|_2^2.
\end{equation}

After training, the model can generate samples from an arbitrary viewpoint given an image $\mathbf I$:
\begin{equation}
\label{eq:0123}
\mathbf I_{\delta\mathbf R}=\mathrm{Zero123}(\mathbf I, \delta\mathbf R).
\end{equation}

Though free-viewpoint data can be generated using the pure 2D pipeline, it lacks rigorous 3D consistency and is unable to control articulated human poses.

\vspace{-0.4cm}
\paragraph{Score distillation sampling.} 
While the diffusion model exhibits remarkable proficiency in generating 2D images, it is constrained by its inherent incapacity to produce a 3D representation directly. For 3D object generation from text prompt, DreamFusion proposes the SDS loss \cite{bib:DreamFusion} that optimized a 3D representation with diffusion guidance, which can be formulated as
\begin{equation}
\label{eq:sds}
\nabla_\eta \mathcal{L}_{SDS} \triangleq \mathbb{E}_{t, \epsilon}\left[w(t)\left(\hat{\epsilon}\left(\mathbf{z}_t, t, y\right)-\epsilon\right) \frac{\partial \mathbf{x}}{\partial \eta}\right],
\end{equation}
where $\eta$ represents the optimization parameters; $w$ is a weighting function that depends on the timestep; and $y$ is the text-conditioned feature.

\subsection{Drivable Tetrahedral Representation}
\label{sec:tet}

Following DMTet \cite{bib:DMTet}, we employ a hybrid representation for 3D geometry. As introduced in Sec~\ref{sec:pre}, this method depends on a pre-defined tetrahedral grid, along with learnable vertex displacements $\delta\mathbf V^t=\{\delta\mathbf v^t\}$ and signed distance values $\mathbf S=\{s\}$, to represent an arbitrary 3D geometry. Based on its volume subdivision method, the triangular mesh $\mathcal M=(\mathbf V^m,\mathbf F)$ can be obtained with vertices $\mathbf V^m=\{\mathbf v^m\}$ and faces $\mathbf F$. 

Nevertheless, the tetrahedral grid can only represent a static scene, while the human body has dynamic articulated poses. To tackle human articulated motion, we introduce the skinning mechanism \cite{bib:SMPL} for the generated mesh. As for each vertex $\mathbf v^m$, we find the nearest triangle on a parametric mesh \cite{bib:SMPLX,bib:MANO} with vertices $\{\mathbf v^p_1,\mathbf v^p_2,\mathbf v^p_3\}$. The skinning weights and expression blendshapes of $\mathbf v^m$ can be formulated as follows,
\begin{equation}
\label{eq:w}
\begin{array}{ll}
     & \mathbf W_{\mathbf v^m} = u\mathbf W^p_{\mathbf v^p_1} + v\mathbf W^p_{\mathbf v^p_2} + \gamma\mathbf W^p_{\mathbf v^p_3} \\[5pt]
     & \mathbf E_{\mathbf v^m} = u\mathbf E^p_{\mathbf v^p_1} + v\mathbf E^p_{\mathbf v^p_2} + \gamma\mathbf E^p_{\mathbf v^p_3},
\end{array}
\end{equation}
where $u,v,\gamma$ denote the barycentric coordinates of the projection of  $\mathbf v^m$ onto the face; $\mathbf W^p,\mathbf E^p$ are the skinning weights and blendshapes defined by SMPLX \cite{bib:SMPLX} or MANO \cite{bib:MANO}. With the retrieval of $\mathbf W,\mathbf E$, mesh vertices can be deformed to the posed space:
\begin{equation}
\label{eq:skin}
    \tilde{\mathbf v}^m=\sum_{b=1}^B \mathbf{W}_{\mathbf v^m, b} \mathbf G_b(\boldsymbol\theta, \mathbf{J}) \mathbf G(\mathbf{0}, \mathbf{J})^{-1} (\mathbf v^m+\mathbf E_{\mathbf v^m}\boldsymbol{\psi}),
\vspace{-0.1cm}
\end{equation}
where $\mathbf G$ is the kinematic transformation matrix, $b$ indexes articulated bones, and $\mathbf J$ denotes bone joints. Please refer to SMPL \cite{bib:SMPL} for more details of the skinning mechanism.

Considering the limited quantity of training data and various articulated poses, the initialization of the tetrahedral grid is important yet non-trivial. Benefiting from the aforementioned drivable mechanism, we can use a canonical SMPLX template mesh for the initialization, as shown in Fig.~\ref{fig:arch}. This approach allows us to incorporate human geometry prior into the 3D representation.

In addition, a texture field is adopted for colored appearance. Following Get3D \cite{bib:Get3D}, we find the surface points $\mathbf P_s$ on $\mathcal M$ that align with pixels by the rasterization of the deformed mesh $(\{\tilde{\mathbf v}^m\}, \mathbf F)$. Then, the texture field $\mathcal C$ with MLPs can predict RGB values $\hat{\mathbf I}$ as follows,
\begin{equation}
\label{eq:color}
\mathcal C(\mathbf P_s): \mathbf P_s\in\mathbb R^{H\times W\times 3} \rightarrow \hat{\mathbf I}\in\mathbb R^{H\times W\times 3}.
\end{equation}

\subsection{Optimization}
\label{sec:opt}

We use few-shot images $\{\mathbf I_i\}_{i=1}^N$ to optimize the drivable tetrahedral representation with the optimization parameter $\eta_{\mathcal G}=\{\delta\mathbf V^t,\mathbf S,\mathcal C\}$. For each image, we use off-the-shelf tools \cite{bib:SMPLX,bib:mobrecon} to obtain parametric geometry $\{\boldsymbol\theta_i,\boldsymbol\psi_i\}_{i=1}^N$ with pose and expression coefficients. Given aligned or novel viewpoints $\mathbf R^{align}$, $\mathbf R^{novel}$, images can be rendered with our model:
\begin{equation}
\label{eq:render}
\begin{array}{ll}
     & \hat{\mathbf I}_i^{align} = \mathcal G(\mathbf R^{align}_i,\boldsymbol\theta_i,\boldsymbol\psi_i) \\[5pt]
     & \hat{\mathbf I}_i^{novel} = \mathcal G(\mathbf R^{novel},\boldsymbol\theta_i,\boldsymbol\psi_i).
\end{array}
\end{equation}

\paragraph{Few-shot reference for body.}
In order to enhance the ability to express human body characteristics with few-shot images $\{\mathbf I_i\}_{i=1}^N$, we use off-the-shelf tools \cite{bib:ICON,bib:omnidata} to estimate the mask $\{\mathbf M_i\}_{i=1}^N$, normal $\{\mathbf N_i\}_{i=1}^N$, and depth $\{\mathbf D_i\}_{i=1}^N$. Then, we design reconstruction losses as follows,
\begin{equation}
\label{eq:recon}
\begin{array}{llll}
     & \mathcal L_{texture} = \mathrm{LPIPS}(\hat{\mathbf I}_i^{align},\mathbf I_i) + ||\hat{\mathbf I}_i^{align}-\mathbf I_i||_2^2 \\[5pt]
     & \mathcal L_{normal} = 1- <\hat{\mathbf N}_i^{align},\mathbf N_i> \\[5pt]
     & \mathcal L_{depth} = \frac{\mathrm{cov}(\hat{\mathbf D}_i^{align},\mathbf D_i)}{\sigma_{\hat{\mathbf D}_i^{align}}\sigma_{\mathbf D_i}} \\
     & \mathcal L_{mask} = ||\hat{\mathbf M}_i^{align} - {\mathbf M}_i||_2^2,
\end{array}
\end{equation}
where $<\cdot,\cdot>$ represents cosine similarity, $\mathrm{LPIPS}$ computes perceptual similarity \cite{bib:LPIPS}. $\mathcal L_{depth}$ is formulated as pearson correlation coefficient \cite{bib:peason}.

Therefore, $\mathcal L_{recon} = \mathcal L_{texture} + \mathcal L_{normal} + \mathcal L_{depth} + \mathcal L_{mask}$ for body avatar reconstruction.
In addition, we render high-resolution (\ie, $256\times 256$) hand/head regions to compute $\mathcal{L}_{recon}^{hand},\mathcal{L}_{recon}^{head}$ using Eq.~(\ref{eq:recon}) and add them to the final $\mathcal L_{recon}$.

\vspace{-0.4cm}
\paragraph{Few-shot reference for hand.}
The properties of the hand are distinctive from the body, \ie, subtle wrinkles and bare geometry. Hence, the benefits from normal/depth supervision are limited and we only use $\mathcal L_{texture}$ and $\mathcal L_{mask}$ in Eq.~(\ref{eq:recon}) for hand optimization. In addition, we design a Laplacian constraint for geometry smoothness as follows,
\begin{equation}
\label{eq:lap}
\mathcal L_{lap} = ||\mathbf L_{\mathcal M}\cdot\hat{\mathbf{\mathbf{n}}}_{\mathcal M}||_2,
\end{equation}
where $\mathbf L_{\mathcal M}$ and $\hat{\mathbf{\mathbf{n}}}_{\mathcal M}$ denote Laplacian matrix and vertex normals for the generated triangle mesh $\mathcal M$. Finally, $\mathcal L_{recon}=\mathcal L_{texture}+\mathcal L_{mask}+\lambda_{lap}\mathcal L_{lap}$ for hand avatar creation.

\vspace{-0.4cm}
\paragraph{Few-shot guidance.}
Few-shot images cannot cover complete visual features of the human body because of sparse viewpoints and articulated self-occlusion. Therefore, we improve random-view rendering with the diffusion prior.

Specifically, pre-trained Zero123 \cite{bib:Zero123} is employed as the prior model, which takes image features and viewpoints as the condition to produce random-viewpoint images, as described in Sec.~\ref{sec:pre}. Given $\hat{\mathbf I}_i^{novel}$, we first use Eq.~(\ref{eq:forward}) to sample nosing image $\mathbf z_{i,t}$ with gaussian noise $\epsilon$ and compute relative viewpoint $\delta \mathbf R$ based on $\mathbf R^{align}_i$ and $\mathbf R^{novel}$. To generate gradients to optimize the 3D representation, we employ SDS loss derived by Zero123 as follows,
\begin{equation}
\label{eq:0123sds}
\nabla_{\eta_{\mathcal G}} \mathcal{L}_{sds} = \mathbb{E}_{t, \epsilon}\left[w(t)\left(\hat{\epsilon}\left(\mathbf{z}_{i,t}, t,\mathrm{CLIP}(\mathbf I_i), \delta\mathbf R\right)-\epsilon\right) \frac{\partial \mathbf{x}}{\partial \eta_{_{\mathcal G}}}\right].
\end{equation}

Overall, $\mathcal L=\mathcal L_{recon}+\lambda_{sds}\mathcal L_{sds}$ is employed to optimize the 3D representation $\mathcal G$ for creating human avatars.

\section{Experiments}
\label{sec:experiments}

\subsection{Datasets}

We build new dataset benchmarks to investigate the task of few-shot dynamic human reconstruction. The purpose of our dataset is to generate training data with casual human poses and conduct multi-view evaluations under the canonical human pose. To present sufficient information with few-shot data, extremely articulated self-occlusion is neglected.

For quantitative metrics, we report LPIPS \cite{bib:LPIPS}, PSNR, and SSIM \cite{bib:ssim} to reflect rendering quality.

\vspace{-0.4cm}
\paragraph{FS-XHumans.} 
XHumans \cite{bib:XAvatar} offer 3D clothed human scans with 20 different identities and various poses. For each identity, we choose 8 scans with different poses and render them from different viewpoints to create our training data. Furthermore, as XHumans do not provide canonical-pose data, we select the scan from their dataset that most closely matches the A-pose and render it from 24 spherically distributed viewpoints for evaluation.

\vspace{-0.4cm}
\paragraph{FS-DART.}
We utilize DART \cite{bib:DART}, a hand texture model, to generate 100 hand identities with distinct shapes and textures. To acquire training poses, we collect real hand poses using a monocular reconstruction method \cite{bib:mobrecon}. Each hand identity has 8 training data with varying poses and viewpoints. For evaluation, we render zero-pose hand samples from 24 spherically distributed viewpoints.

Please refer to \textit{suppl. material} for more dataset details.


\subsection{Ablation Studies}
\begin{figure}[t]
\begin{center}
\includegraphics[width=\linewidth]{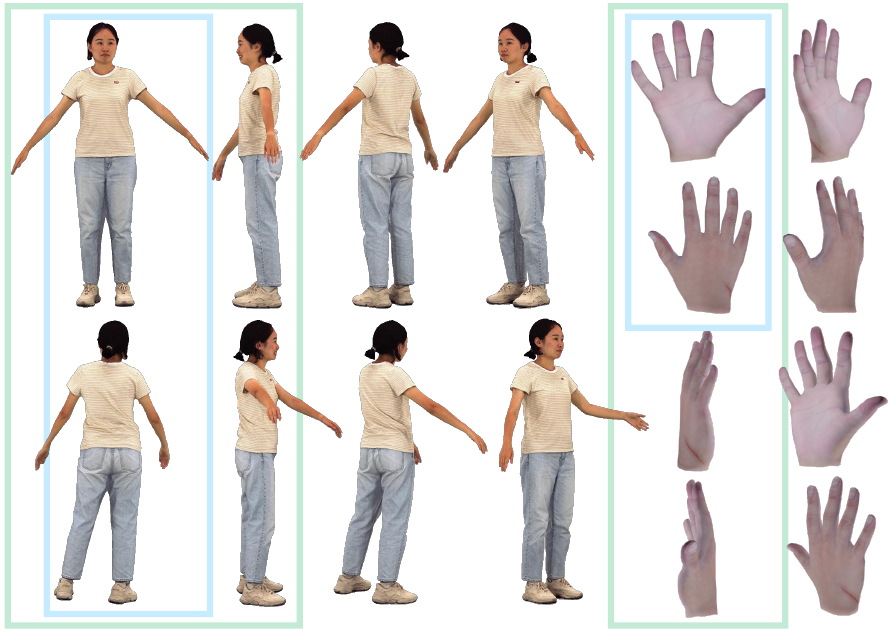}
\caption{Training data for ablation studies in Figs.~\ref{fig:multibody} and \ref{fig:multihand}. Blue and green boxes indicate 2- and 4-shot training data, respectively.
}
\label{fig:data}
\end{center}
\vspace{-0.3cm}
\end{figure}
\begin{table}[t]
\small
\renewcommand{\arraystretch}{1.1}
\centering
\begin{tabular}{c | c c c }
\hline
Method & PSNR $\uparrow$ & SSIM $\uparrow$ & LPIPS $\downarrow$ \\
\hline 
& \multicolumn{3}{c}{\textit{4-shot FS-XHumans}} \\
$\lambda_{sds}=0$ & 24.54 & 0.9522 & 0.0379\\
$\lambda_{sds}=0.01$ & \bf 25.64 & \bf 0.9627 & \bf 0.0347\\
$\lambda_{sds}=0.05$ & 25.10 & 0.9562 & 0.0373\\
$\lambda_{sds}=1$ & 24.56 & 0.9413 & 0.0403\\
\hline
\hline
& \multicolumn{3}{c}{\textit{2-shot FS-DART}} \\
$\lambda_{sds}=0, \lambda_{lap}=1$ & 25.62 & 0.9532 & 0.0627\\ 
$\lambda_{sds}=0.01, \lambda_{lap}=1$ & 26.22 & 0.9635 & \bf 0.0559 \\
$\lambda_{sds}=0.05, \lambda_{lap}=1$ & \bf 26.61 & \bf 0.9673 & 0.0577 \\
$\lambda_{sds}=1, \lambda_{lap}=1$ & 26.01 & 0.9656 & 0.0736\\
$\lambda_{sds}=0.05,\lambda_{lap}=0$ & 26.13 & 0.9648 & 0.0725\\
$\lambda_{sds}=0.05,\lambda_{lap}=5$ & 25.69 & 0.9633 & 0.0602\\
\hline
\end{tabular}
\caption{The effects of SDS and Laplacian normal losses.}
\label{tab:abl}
\vspace{-0.1cm}
\end{table}
\begin{figure}[t]
\begin{center}
\includegraphics[width=\linewidth]{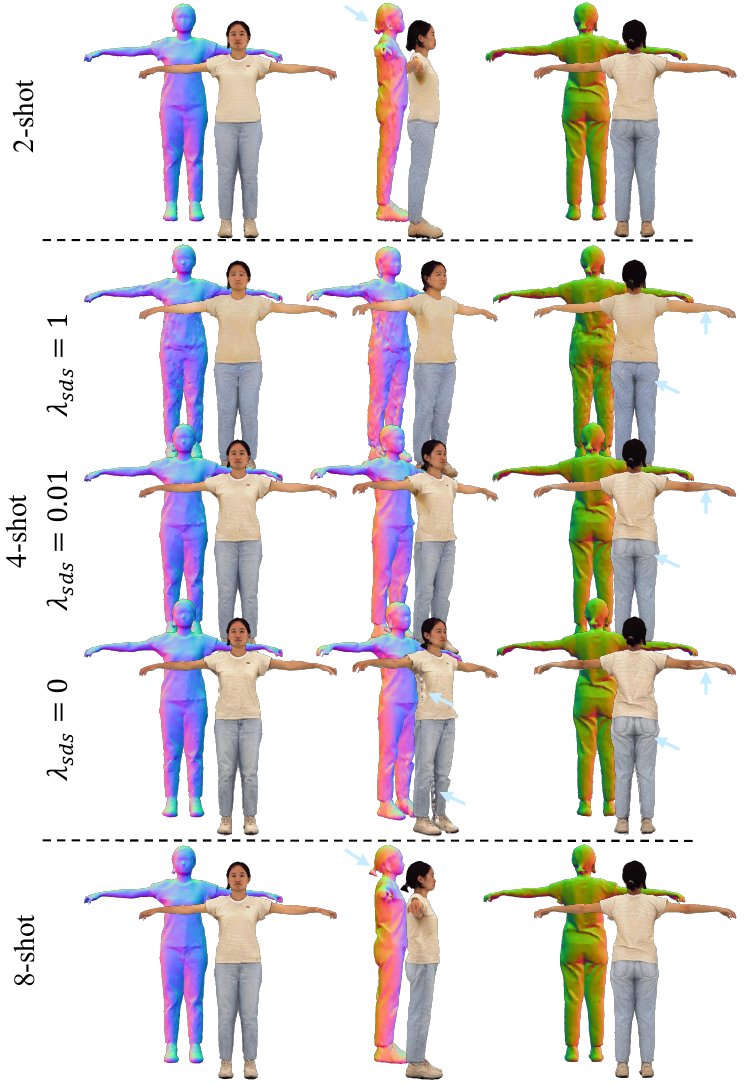}
\caption{Ablation studies on the few-shot body reconstruction task. Zoom in to see details.
}
\label{fig:multibody}
\end{center}
\vspace{-0.6cm}
\end{figure}

\begin{figure}[t]
\begin{center}
\includegraphics[width=0.95\linewidth]{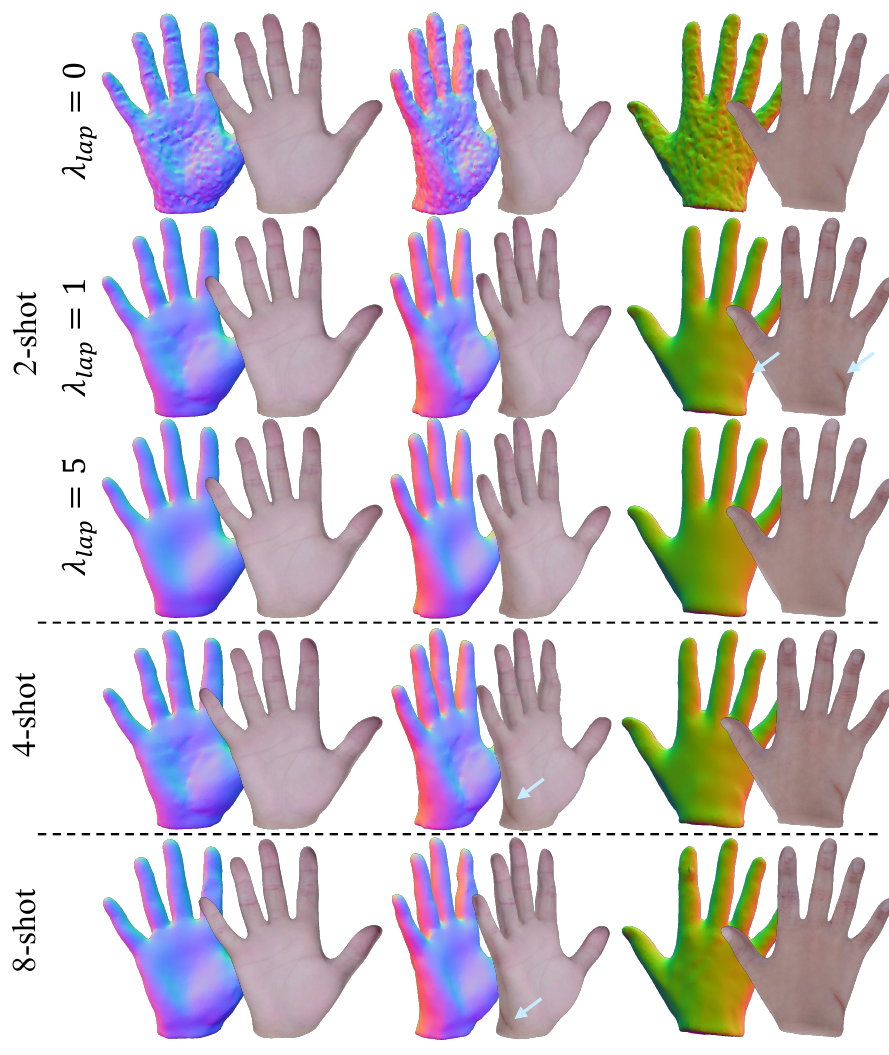}
\caption{Ablation studies on the few-shot hand reconstruction task. Zoom in to see details.
}
\label{fig:multihand}
\end{center}
\vspace{-0.4cm}
\end{figure}

Considering that the primary information can be presented on the palm and back of the hand, we use the 2-shot setting for hand ablation studies. In contrast, the lateral body holds important information, so we employ the 4-shot setting to study the performance on the human body. Finally, we report our $N$-shot ($N=2,4,8$) results.
The training data used for Figs.~\ref{fig:multibody} and \ref{fig:multihand} are shown in Fig.~\ref{fig:data}. 

\vspace{-0.4cm}
\paragraph{Effect of few-shot guidance.}
The SDS loss is utilized to facilitate the reconstruction of unseen regions. Therefore, the effect of SDS should be compatible with unseen areas. That is, in contrast to the 4-shot setting, the 2-shot reconstruction exhibits a heightened reliance on the SDS loss due to the lack of information. As shown in Table~\ref{tab:abl}, $\lambda_{sds}$ offers different optimal choices for varying training data amounts.

The effect of SDS loss is also demonstrated in Fig.~\ref{fig:multibody}. As indicated by arrows, $\lambda_{sds}=0$ prevents the model from presenting reasonable unseen textures. Besides, over-size SDS constraints would lead to the problem of color distortion, thereby harming the reconstruction quality. As a result, we use $\lambda_{sds}=0.05,0.01,0.01$ for 2-, 4-, and 8-shot reconstruction settings.

\vspace{-0.4cm}
\paragraph{Effect of Laplacian normal constraint.}
Instead of ground-truth normal/depth supervision, we design a Laplacian normal loss to regularize hand geometry. As illustrated in Fig.~\ref{fig:multihand}, $\lambda_{lap}=0$ induces fractured geometry, and $\lambda_{lap}=5$ can produce a over-smooth geometry. These two situations are not suited to represent the hand. That is, the hand has subtle palm wrinkles and texture. If the geometry can effectively capture wrinkles, it could contribute to generating faithful textures. Surprisingly, our experiment shows that the proper Laplacian normal constraint can achieve this. As shown, $\lambda_{lap}=1$ promotes our model to represent detailed hand shapes. Benefiting from the informative geometry, the texture reconstruction is more accurate, as indicated by arrows.

\vspace{-0.4cm}
\paragraph{Ablation study on $N$-shot reconstruction.}

Our framework can reconstruct a human avatar with an arbitrary quantity of images. We argue that one-shot data cannot furnish adequate information for human reconstruction, so we demonstrate 2-, 4-, and 8-shot tasks as examples. Referring to our results in Table~\ref{tab:comp}, 2-shot metrics are very close to those of the 8-shot task, indicating our approach is adept at reconstructing humans from minimal data amount. When it comes to the hand reconstruction task, some 2-shot results are even better than those of the 8-shot task. The primary reason is that the hand feature is predominantly influenced by the hand palm or back, with the lateral hand contributing only minimal additional information.

Similar conclusions can be observed in Figs.~\ref{fig:multibody} and \ref{fig:multihand}. Visually, the 8-shot task shows the best quality with faithful hair geometries, cloth textures, palm wrinkles, \etc.
\begin{table}[t]
\small
\renewcommand{\arraystretch}{1.1}
\setlength\tabcolsep{3pt}
\centering
\begin{tabular}{c | c c c }
\hline
Method & PSNR $\uparrow$ & SSIM $\times 10^{2} \uparrow$ & LPIPS $\times 10^{2} \downarrow$ \\
\hline 
& \multicolumn{3}{c}{\textit{FS-XHumans}} \\
SelfRecon &  19.9/20.7 & 92.7/94.3 & 6.5/6.3 \\
TeCH & 21.0 & 92.4 & 6.5 \\
HaveFun (ours) & 24.0/25.6/\textbf{26.8} & 95.5/96.3/\textbf{96.7} & 4.2/3.5/\textbf{3.0} \\
\hline
& \multicolumn{3}{c}{\textit{FS-DART}} \\
SelfRecon & 20.8/21.3/21.7 & 92.0/92.4/92.8 & 9.6/9.0/8.7 \\
HaveFun (ours) & 26.6/\textbf{26.7}/26.3 & 96.7/\textbf{96.8}/96.7 & 5.8/5.5/\textbf{5.2}\\
\hline
\end{tabular}
\caption{Comparison of few-shot human reconstruction. For FS-Xhumans, TeCH \cite{bib:TeCH} only supports single-image reconstruction. SelfRecon \cite{bib:SelfRecon} results follow the format of ``8-/100-shot'', whereas our metrics are provided for ``2-/4-/8-shot'' tasks. For FS-DART, all metrics are for ``2-/4-/8-shot'' tasks.}
\label{tab:comp}
\vspace{-0.4cm}
\end{table}

\begin{figure}[t]
\begin{center}
\includegraphics[width=0.93\linewidth]{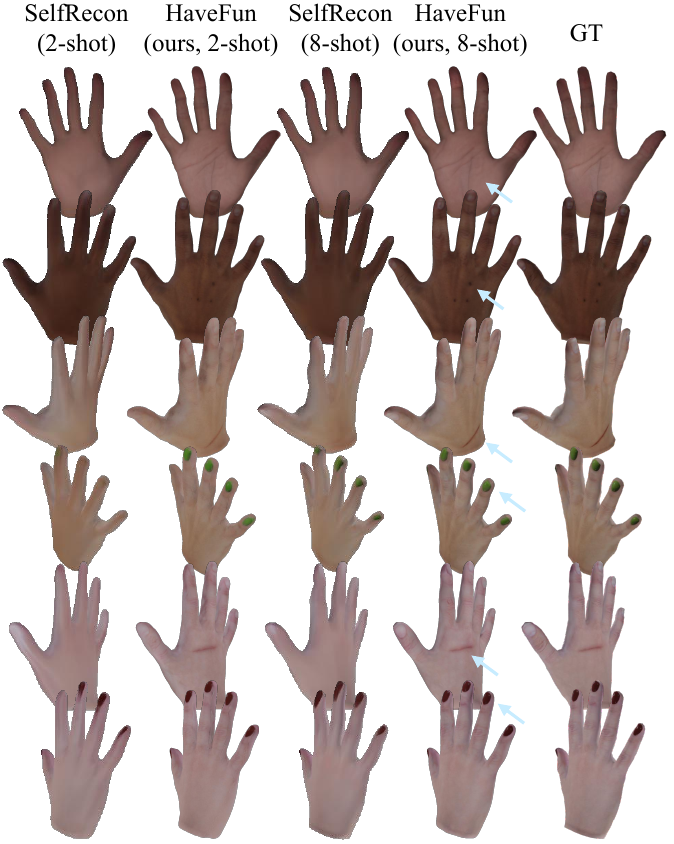}
\caption{Comparison of hand reconstruction on FS-DART. See \textit{suppl. material} for training data.
}
\label{fig:handcomp}
\end{center}
\vspace{-0.7cm}
\end{figure}

\begin{figure*}[t]
\begin{center}
\includegraphics[width=0.95\linewidth]{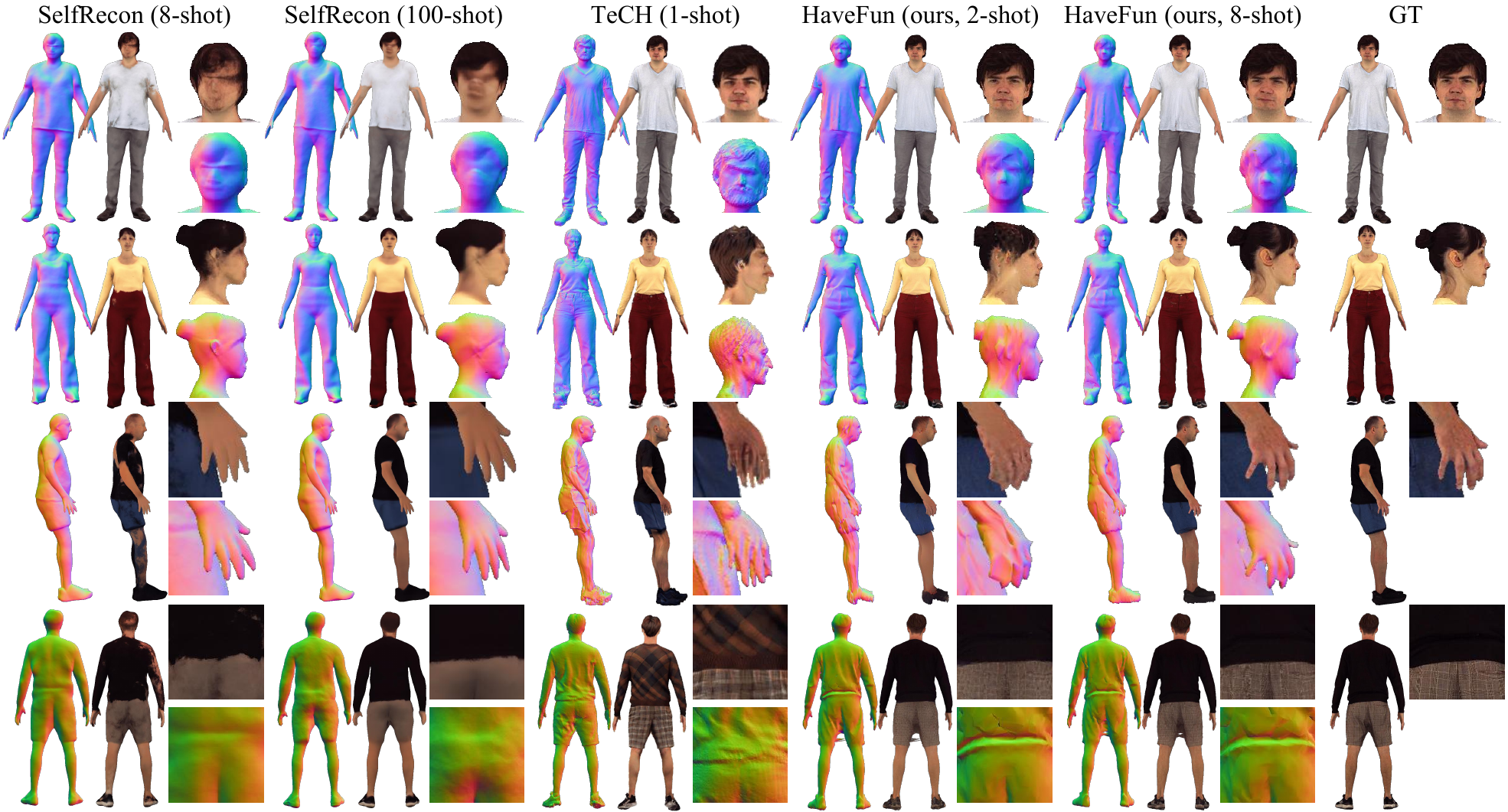}
\caption{Comparison of body reconstruction on FS-XHumans, where TeCH is a 1-shot method and SelfRecon is illustrated with 8-shot or video (100-shot) data training. See \textit{suppl. material} for training data.
}
\label{fig:bodycomp}
\end{center}
\vspace{-0.5cm}
\end{figure*}

\begin{figure}[t]
\begin{center}
\includegraphics[width=0.95\linewidth]{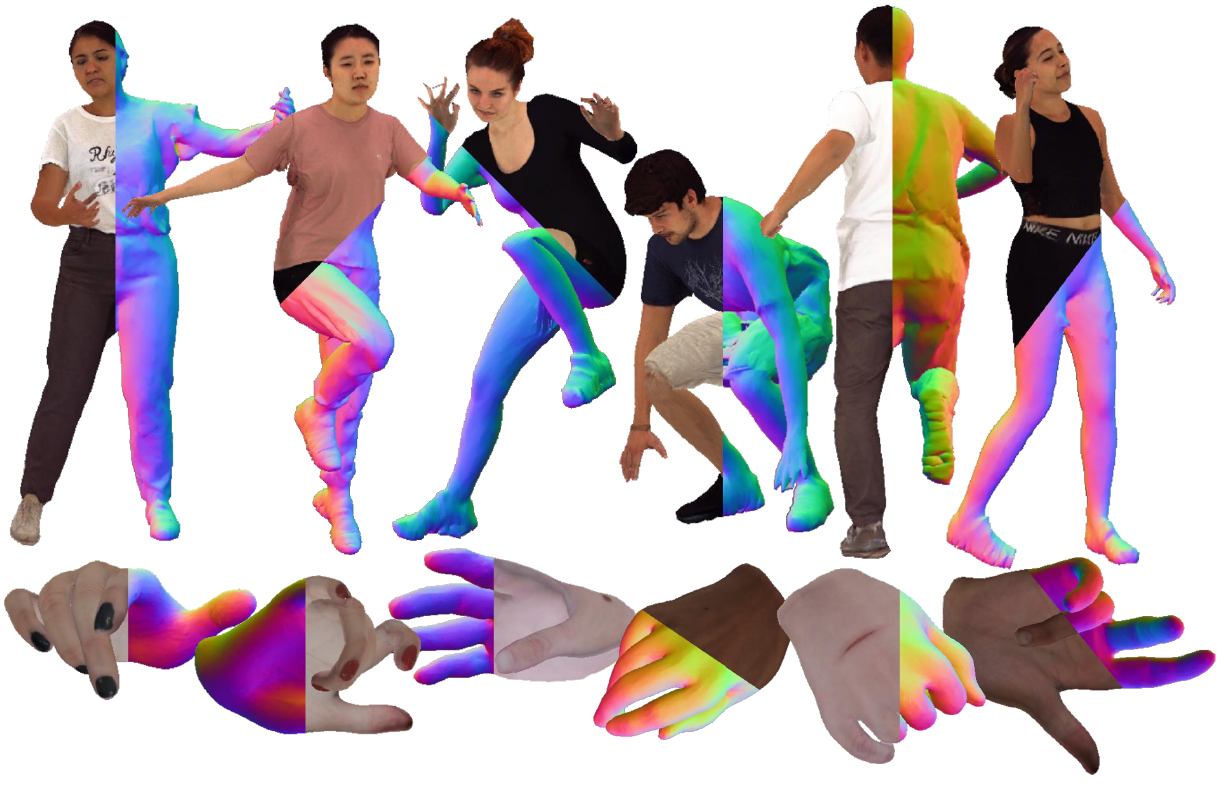}
\caption{Articulated animation of human avatars.
}
\label{fig:deform}
\end{center}
\vspace{-0.7cm}
\end{figure}

\subsection{Comparison with Prior Arts}

We compare our approach with SelfRecon \cite{bib:SelfRecon} and TeCH \cite{bib:TeCH}. The former can take dynamic images or videos as the input, whereas the latter uses the DMTet and SDS prior for static human reconstruction. We present typical results in Figs.~\ref{fig:handcomp} and \ref{fig:bodycomp}. As shown, SelfRecon tends to generate overly smooth human appearances and inherently fails to learn effectively from limited data. Moreover, due to the absence of expression handling, SelfRecon still struggles to produce a plausible portrait, even when trained using video (100-shot) data.
Text-based TeCH can produce detailed texture and sound geometry with a single image. However, TeCH cannot perform a faithful reconstruction. As shown in the first row of Fig.~\ref{fig:bodycomp}, facial identity cannot preserved by TeCH. Furthermore, features from the text caption would be improperly introduced to the avatar. Referring to the second row of Fig.~\ref{fig:bodycomp}, the front head is aligned with the reference image while the lateral head is dominated by the BLIP \cite{bib:BLIP} caption of ``caucasian''. Hence, since the caption cannot perfectly describe visual details, the text-based approach is akin to a generative method rather than a reconstructive one. In contrast, thanks to pure visual prompts from few-shot images, our method can perform a faithful avatar reconstruction for the body, face, and hand. 

It is worthwhile to revisit our 2- and 8-shot results again. As shown in the first row of Fig.~\ref{fig:bodycomp}, the 2-shot result is better than the 8-shot one in terms of portrait reconstruction. This discrepancy arises from the diverse expressions presented in training data. That is, the 2-shot task relies on a single image to describe the face, while the 8-shot task involves lateral body data with various facial expressions. 
Due to the imperfect expression blendshapes from SMPLX, the performance of 8-shot portrait reconstruction from data with diverse expressions is constrained. In addition, the 8-shot reconstruction yields enhanced results for hands and lateral textures owing to the expanded visibility of data.

For quantitative comparison, our method achieves the best results in all metrics, as shown in Table~\ref{tab:comp}, indicating superior rendering quality of the HaveFun framework.

\subsection{Applications}

\paragraph{Animation results.}
Thanks to the drivable tetrahedral representation, our method can perform free-pose articulated deformation for the body and hand so that complex human motion can be presented, as shown in Fig.~\ref{fig:deform}.

\begin{figure}[t]
\begin{center}
\includegraphics[width=\linewidth]{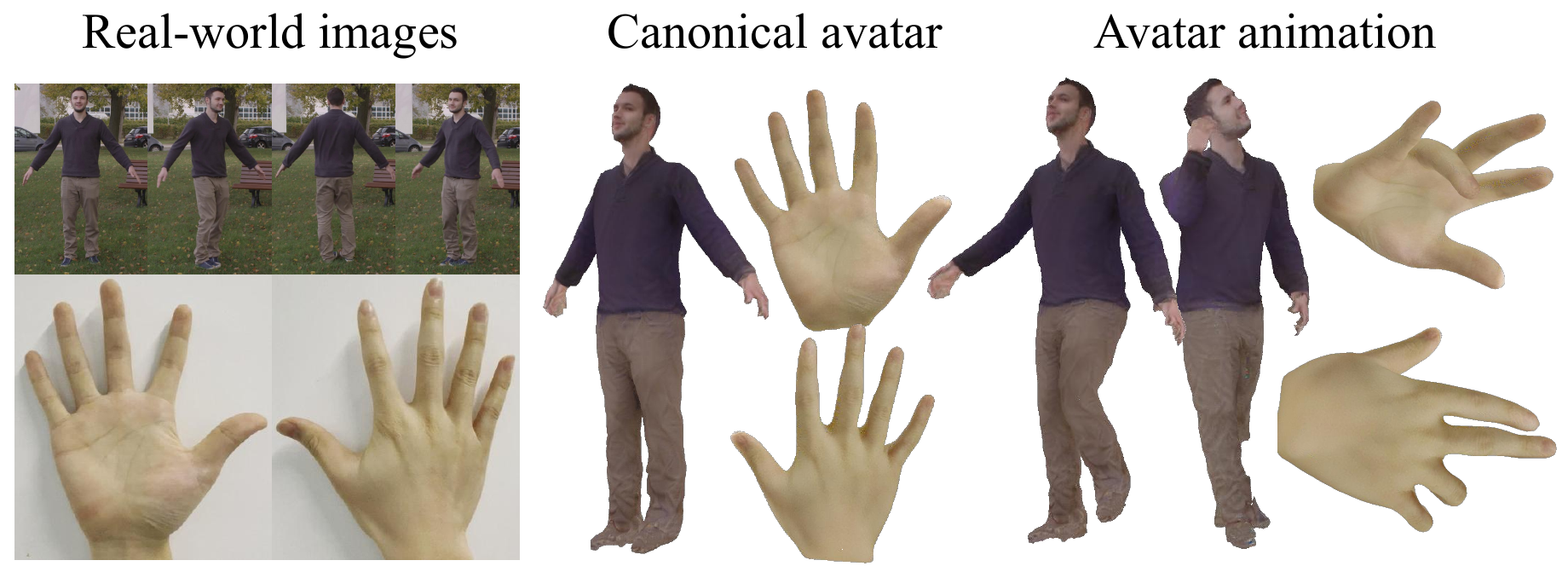}
\caption{Avatar reconstruction from real-world captured data. 
}
\label{fig:realhand}
\end{center}
\vspace{-0.5cm}
\end{figure}

\vspace{-0.4cm}
\paragraph{Reconstructing human avatars with real-world casual capture.} The FS-DART is a synthetic dataset, and the FS-XHumans provides real human images but captured in a studio. Therefore, we acquire real-world unconstrained images to validate our approach. As shown in Fig.~\ref{fig:realhand}, based on 2 or 4 images, our method excels in canonical-space avatar reconstruction and free-pose animation.

\section{Conclusions}
This paper poses a novel research problem of human avatar reconstruction from few-shot unconstrained images. We propose a HaveFun framework with a drivable tetrahedral representation to solve this issue. To optimize our 3D representation, we design a two-phase method with few-shot reference and few-shot guidance. In addition, we develop evaluation benchmarks for the human body and hand. As a result, our approach can produce animatable human avatars with superior rendering quality, which we believe enables a new way for real-world avatar creation.


{
\small
\noindent \textbf{Acknowledgment}
The work was supported in part by the Basic Research Project No.HZQB-KCZYZ-2021067 of Hetao Shenzhen-HK S\&T Cooperation Zone, Guangdong Provincial Outstanding Youth Project No. 2023B1515020055, the National Key R\&D Program of China with grant No.2018YFB1800800, by Shenzhen Outstanding Talents Training Fund 202002, by Guangdong Research Projects No.2017ZT07X152 and No.2019CX01X104, by Key Area R\&D Program of Guangdong Province (Grant No.2018B030338001), by the Guangdong Provincial Key Laboratory of Future Networks of Intelligence (Grant No.2022B1212010001), and by Shenzhen Key Laboratory of Big Data and Artificial Intelligence (Grant No.ZDSYS201707251409055). It is also partly supported by NSFC-62172348, Shenzhen General Project No. JCYJ20220530143604010 and China National Postdoctoral Program for Innovative Talents No. BX2023004. 
}

{
    \small

}

\clearpage
\setcounter{page}{1}
\maketitlesupplementary

\section{Dataset Details}

\begin{figure}[t]
\begin{center}
\includegraphics[width=\linewidth]{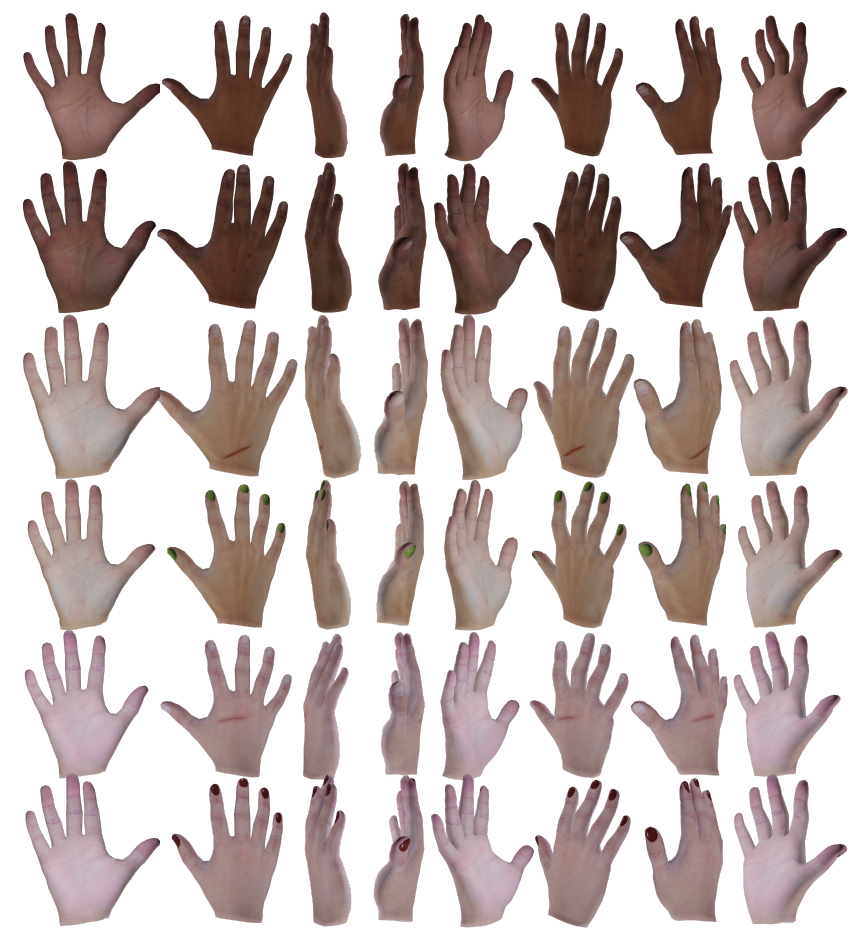}
\caption{Training data for Fig.~6 of the main text. The first $N$ samples in rows are used for the $N$-shot task.
}
\label{fig:handtrain}
\end{center}
\vspace{-0.3cm}
\end{figure}

\paragraph{FS-DART.} Our FS-DART is a synthetic dataset based on the DART \cite{bib:DART} hand model. We create 100 hand identities with a variety of skin colors and hand shapes. In addition, special hand features such as scars, moles, and nail polish are also included in hand textures. As for hand poses, we capture real hand videos and extract pose parameters with MobRecon \cite{bib:mobrecon} for training data creation. As shown in Fig.~\ref{fig:handtrain}, our training samples contain unconstrained casual hand poses. Note that self-occluded poses are not involved so that few-shot data can exhibit sufficient information for the hand reconstruction task. 

\begin{figure}[t]
\begin{center}
\includegraphics[width=0.98\linewidth]{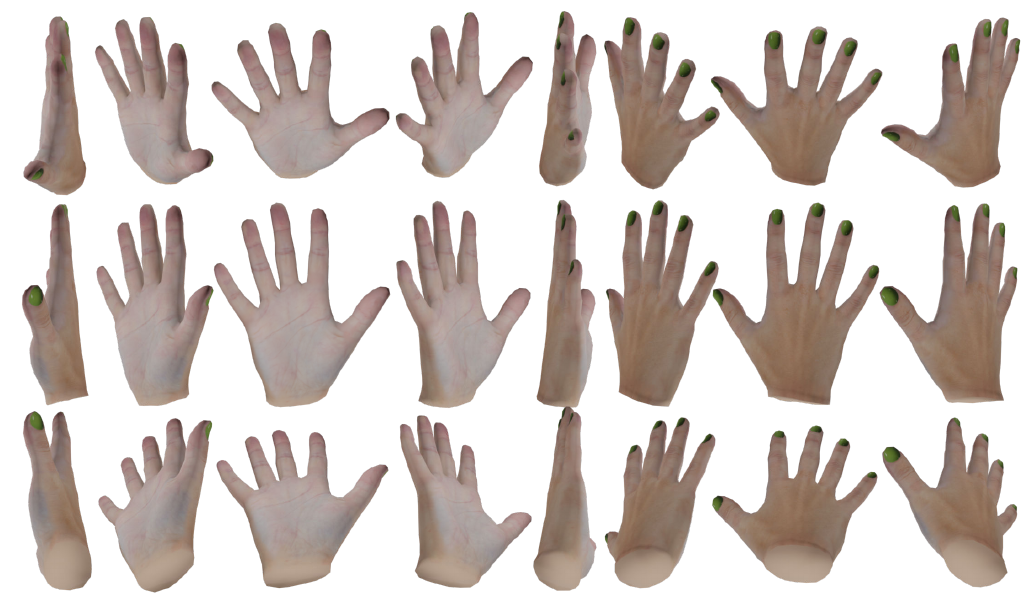}
\caption{Evaluation data of FS-DART from 24 viewpoints.
}
\label{fig:handtest}
\end{center}
\vspace{-0.3cm}
\end{figure}

In terms of evaluation, we assess the effectiveness of our method under the zero hand pose to unveil the performance in shape and texture reconstruction. 
Referring to Fig.~\ref{fig:handtest}, the hand in the zero pose is rendered from 24 sphere-distributed viewpoints, and our model can generate corresponding results for metric computation.

\begin{figure}[t]
\begin{center}
\includegraphics[width=\linewidth]{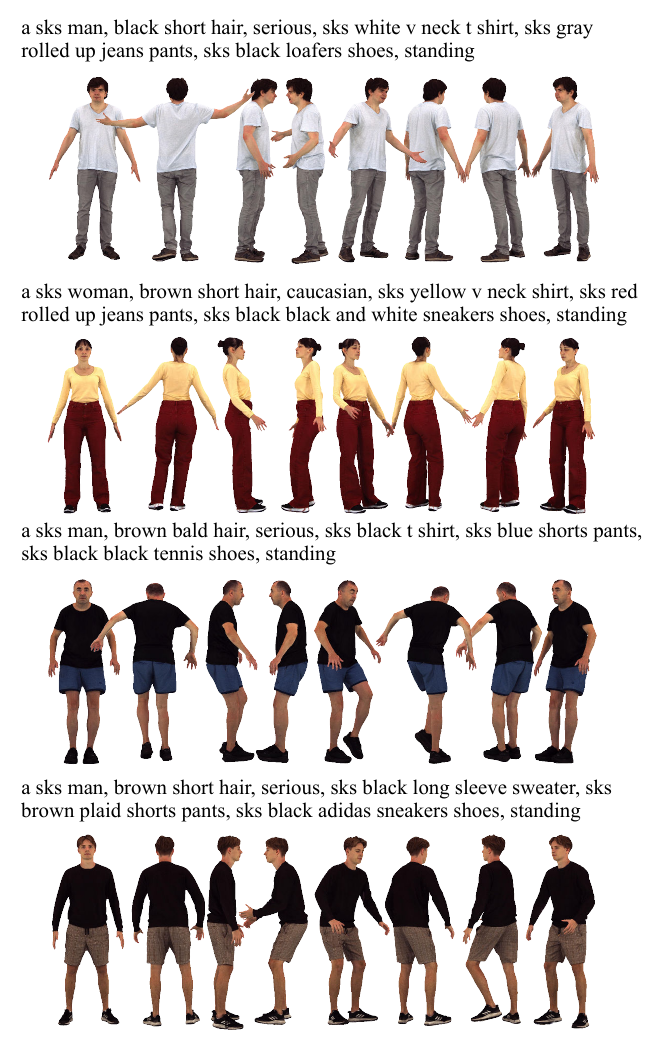}
\caption{Training data for Fig.~7 of the main text. The first $N$ samples in rows are used for the $N$-shot task. The textural captions are only employed by TeCH.
}
\label{fig:bodytrain}
\end{center}
\vspace{-0.5cm}
\end{figure}

\begin{figure}[t]
\begin{center}
\includegraphics[width=\linewidth]{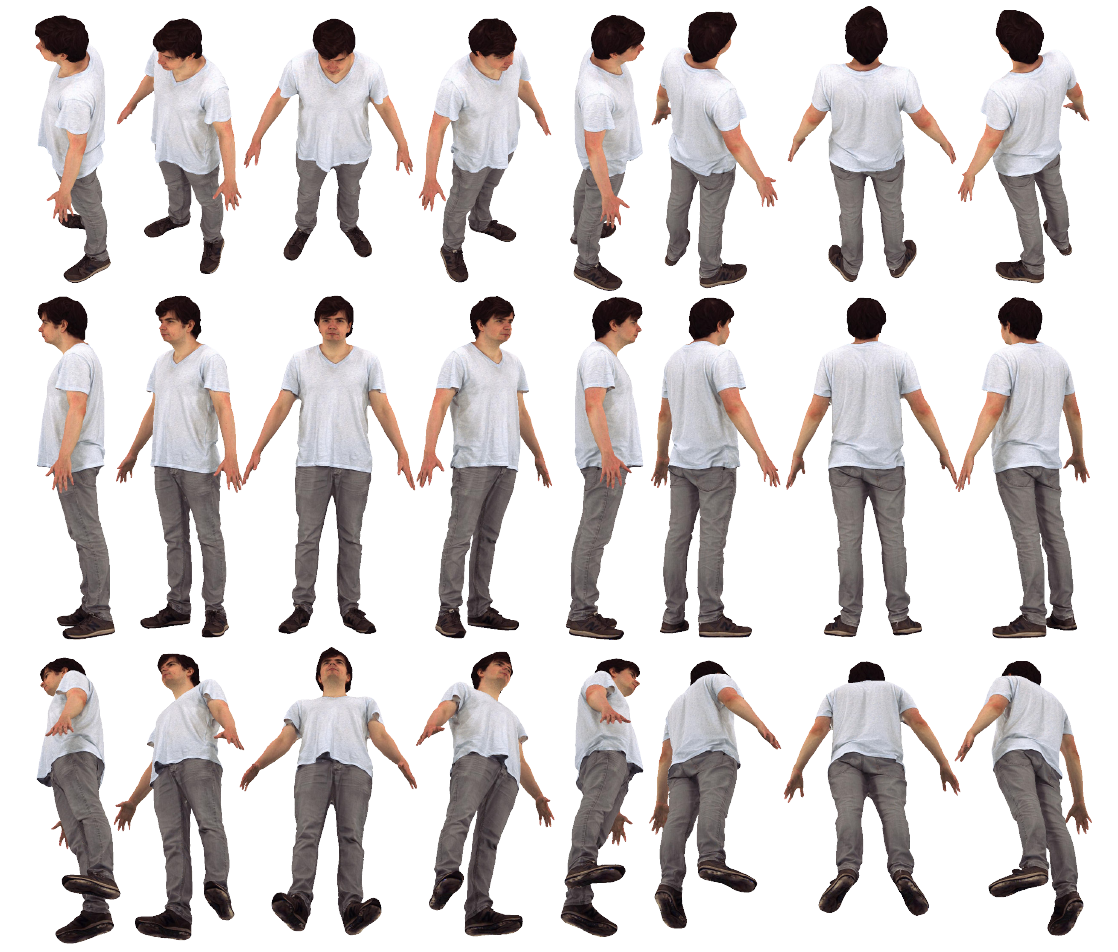}
\caption{Evaluation data of FS-XHumans from 24 viewpoints.
}
\label{fig:bodytest}
\end{center}
\vspace{-0.3cm}
\end{figure}

\vspace{-0.4cm}
\paragraph{FS-XHumans.} Our FS-XHumans dataset is built on really captured XHumans \cite{bib:XAvatar}, which is a 3D scan dataset with 19 actual human identities. For each individual, the XHumans provide 3D scans of motion sequences, including diverse body poses, hand gestures, and facial expressions. Thereby, we select 8 scans from a sequence to produce training data. During data selection, we ensure the diversity of poses and expressions for our training samples, as illustrated in Fig~\ref{fig:bodytrain}. Due to the absence of canonical-pose samples, we opt for a scan closely resembling the A-pose to generate testing samples from 24 sphere-distributed viewpoints for metric computation, as depicted in Fig.~\ref{fig:bodytest}.

It is worthwhile to note that the training data do not have to strictly follow viewpoints in Figs.~\ref{fig:handtrain} and \ref{fig:bodytrain}. We use this viewpoint configuration as an example because it is an efficient setting for few-shot data acquisition. The viewpoints of arbitrarily captured data can be obtained through parametric geometry estimation \cite{bib:SMPLX,bib:mobrecon}. From the perspective of real-world applications, our data setup is reasonable because obtaining data similar to Figs.~\ref{fig:handtrain} and \ref{fig:bodytrain} in practical capture scenarios is straightforward.

\section{Implementation Details}

\paragraph{Tetrahedral grid.} We produce a tetrahedral grid in a $128^3$-size cube using 277,410 vertices and 1,524,684 tetrahedra. Positional displacements and an SDF value are attached to vertices, and we explicitly treat them as optimization parameters without resorting to neural networks.

\vspace{-0.4cm}
\paragraph{Texture field.}
To predict RGB values, we design texture filed $\mathcal C$ using a 3-layer MLP network with a hidden dimension of 64 and a hash positional encoding with a maximum resolution of 2048 and 16 resolution levels. Specifically, the triangle mesh $\mathcal M$ extracted from DMtet is deformed to match the posed human space aligned with the training images. Each pixel is mapped onto the deformed mesh surface, represented by its barycentric coordinates. Then, we query points $\mathbf P_s$ on the canonical triangle mesh $\mathcal M$ with the barycentric coordinates, and the rendered image can be obtained with $\hat{\mathbf I}=\mathcal C(\mathbf P_s)$.

\vspace{-0.2cm}
\paragraph{Optimization details.}

Our experiments are conducted on a NVIDIA A100 GPU. The whole framework is trained in an end-to-end manner.

For the body reconstruction task, the optimization comprises 17,000 iterations. The learning rate starts at 0.05 and is decreased by a factor of 0.1 at the 7,500th and 15,000 steps. The optimization process for a human body takes approximately 4 hours.

In terms of the hand reconstruction task, the optimization requires 2,000 iterations with a learning rate of 0.05. The optimization of a hand identity only costs about 10 minutes.

\section{Details of Compared Methods}
Due to the absence of existing methods designed for few-shot dynamic human reconstruction, we compare HaveFun with a video-based approach and a one-shot static pipeline. 

\vspace{-0.4cm}
\paragraph{SelfRecon.}
In contrast to our data configuration, SelfRecon \cite{bib:SelfRecon} is designed for self-rotated video data. Despite this difference, SelfRecon can perform human reconstruction under our data setup. That is, few-shot unconstrained images used in our work can be treated as key frames of a video. Hence, it is reasonable to compare our approach with SelfRecon. To this end, we acquire officially released implementation codes from \url{https://github.com/jby1993/SelfReconCode} and re-implement the part of the dataset for the adaptation of few-shot image input. In addition, we set a batch size of 2 and a training step of 15,000. The training process costs about 12 hours for a human individual. Furthermore, we also train a SelfRecon model following its original data setting. That is, we generate video data consisting of 100 frames, containing uniformly self-rotated body images, as shown in ``SelfRecon (100-shot)'' in Fig.~7 of the main text. The SelfRecon results are also displayed in our \textit{suppl. video}. As shown, the instability in geometry and texture is evident across different viewpoints due to the employed training samples with highly articulated motion and the intrinsic mechanism of viewpoint-dependent color prediction.

For the hand experiment, we integrate MANO articulation into SelfRecon and adopt the same settings as the body experiment.

\vspace{-0.4cm}
\paragraph{TeCH.} 
TeCH is a one-shot human reconstruction method utilizing SDS guidance, similar to the technical pipeline in our HaveFun framework. For comparison, we employ the official implementation from \url{https://github.com/huangyangyi/TeCH}. TeCH requires 5 stages to optimize a human avatar, including VQA caption, DreamBooth fine-tuning, geometry optimization, geometry post-processing, and texture optimization. The captions used for text-guided SDS loss are shown in Fig~\ref{fig:bodytrain}. In addition, we argue that the stage of geometry post-processing is tricky due to the replacement of the hand shape with the SMPLX hand mesh. That is, the hand is reconstructed using SMPLX rather than TeCH. For a fair experimental setup, we omit the geometry post-processing and jointly optimize the complete geometry and texture. All other settings adhere to the original TeCH report, and it takes approximately 6 hours to generate a human avatar.

As the VQA caption of hands is unexplored, we do not include the comparison of TeCH in the hand task.

\section{More Results}
\paragraph{Effects of normal and depth losses.}
\begin{table}[t]
\small
\renewcommand{\arraystretch}{1.1}
\centering
\begin{tabular}{c c | c c c }
\hline
$\mathcal L_{normal}$ & $\mathcal L_{depth}$ & PSNR $\uparrow$ & SSIM $\uparrow$ & LPIPS $\downarrow$ \\
\hline 
& & \multicolumn{3}{c}{\textit{4-shot FS-XHumans}} \\
$\checkmark$ & $\checkmark$ & \bf 25.64 & \bf 0.9627 & \bf 0.0347\\
$\checkmark$ &  & 25.08 & 0.9601 & 0.0352 \\
             & $\checkmark$ & 24.30 & 0.9581 & 0.0404 \\
             &              &23.85& 0.9575 &0.0466\\
\hline
\end{tabular}
\caption{The effects of normal and depth losses.}
\label{tab:loss}
\end{table}
\begin{figure}[t]
\begin{center}
\includegraphics[width=\linewidth]{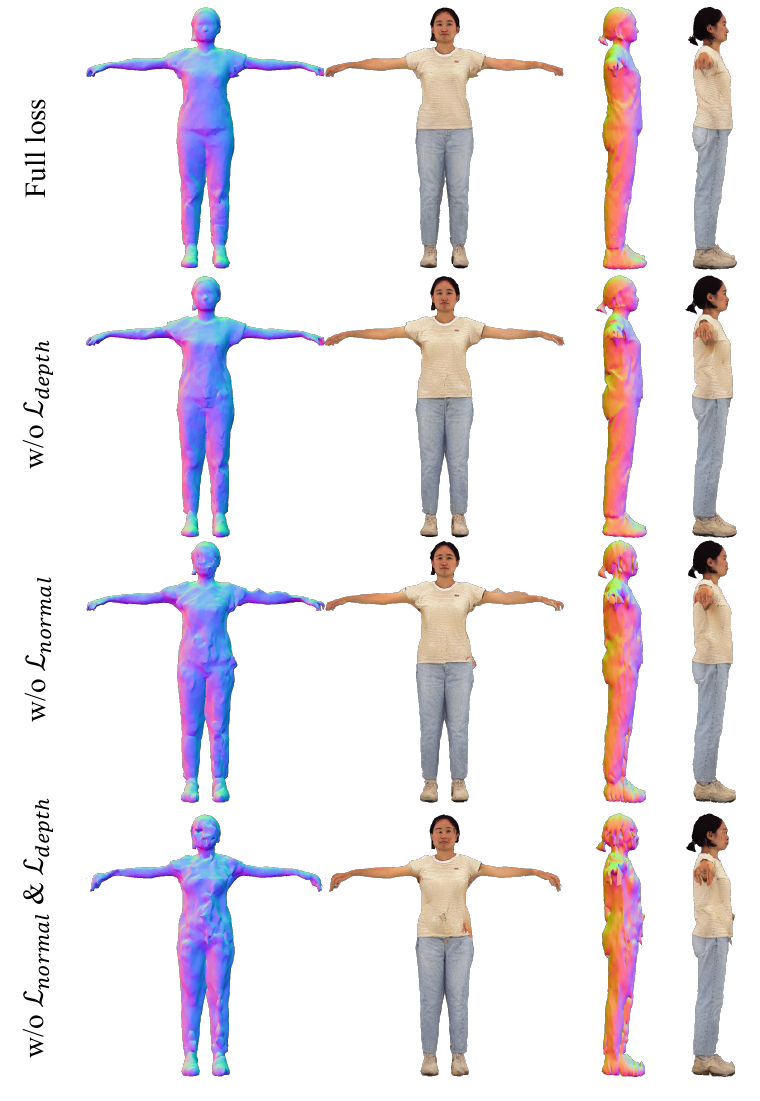}
\caption{The effects of normal and depth losses.
}
\label{fig:loss}
\end{center}
\end{figure}

Referring to Table~\ref{tab:loss} and Fig.~\ref{fig:loss}, normal and depth losses give rise to instructive effects on human avatar reconstruction. Nevertheless, removing depth loss only leads to a minor performance drop.
Due to the often inaccurate estimates of monocular depth, depth supervision is optional in real-world applications, and the HaveFun framework can present human avatars without depth labels.



\vspace{-0.2cm}
\paragraph{SDS loss for the 8-shot task.}
Table~\ref{tab:sds} shows the effect of SDS loss in the 8-shot FS-XHumans experiment, which also supports the conclusion of the main text.

\begin{figure}[t]
\begin{center}
\includegraphics[width=0.95\linewidth]{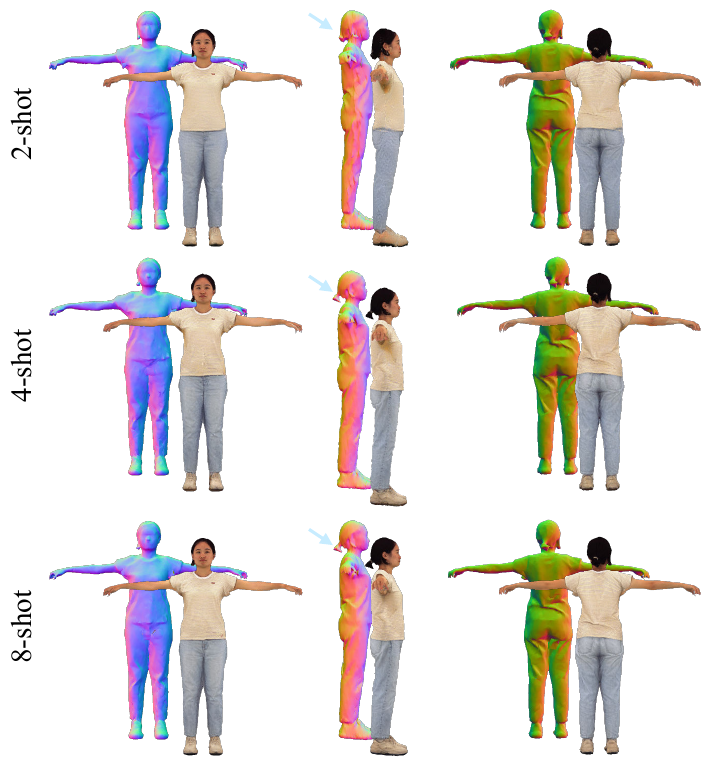}
\caption{Comparison of body reconstruction with few-shot data}
\label{fig:side}
\end{center}
\end{figure}

\vspace{-0.2cm}
\paragraph{Side-view results of the 4-shot setting.}
In Fig.~4 of the main text, we use a side-view for 2/8-shot tasks to highlight the details of hair reconstruction and another view for the 4-shot task to unveil the SDS effect for unseen regions. To fully present these experiments, we supplement 4-shot side-view results in Fig.~\ref{fig:side} for comparison.

\begin{table}[t]
\small
\renewcommand{\arraystretch}{1.1}
\centering
\begin{tabular}{c | c c c }
\hline
Method & PSNR $\uparrow$ & SSIM $\uparrow$ & LPIPS $\downarrow$ \\
\hline 
$\lambda_{sds}=0$ & 26.45 & 0.9604 & 0.0343 \\
$\lambda_{sds}=0.01$ & \bf 26.82 & \bf  0.9674 & \bf 0.0301 \\
$\lambda_{sds}=1$ & 25.40 & 0.9570 & 0.0375 \\
\hline
\end{tabular}
\caption{The SDS effects on the 8-shot FS-XHumans experiment.}
\label{tab:sds}
\end{table}

\begin{figure}[t]
\begin{center}
\includegraphics[width=0.95\linewidth]{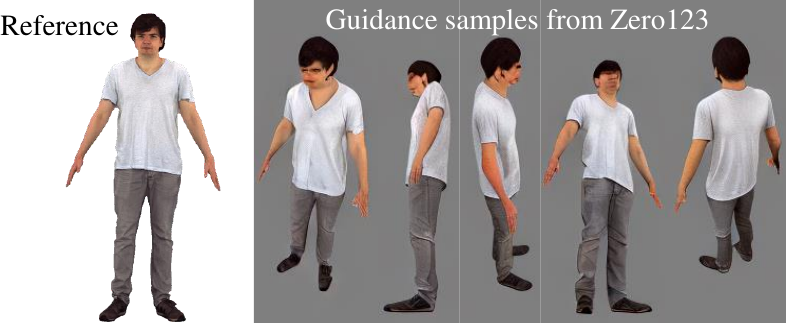}
\caption{Visualization of Zero123 guidance}
\label{fig:0123}
\end{center}
\end{figure}

\paragraph{The Zero123 guidance.}
As shown in the Fig.~\ref{fig:0123}, the purely 2D method Zero123 produces low-quality guidance images (\eg, face). Our model achieves performance beyond Zero123 because of a 3D-aware representation and depth/normal supervision.


\vspace{-0.2cm}
\paragraph{More results in dynamic demonstration.}
Please refer to the project page \url{https://seanchenxy.github.io/HaveFunWeb} for dynamic results.

\section{Limitations and Future Works}
\label{sec:limit}

\paragraph{Expression control.}
To handle varying expressions in training data, we transform expression blendshapes defined by SMPLX \cite{bib:SMPLX} into our framework. Nevertheless, the blendshapes are not accurate enough, harming the precision of expression control. The impact on portrait reconstruction is explained in Fig.~7 of the main text. To tackle this difficulty, we will introduce advanced expression control methods (\eg, \cite{bib:PDFGC}) to the HaveFun framework. 

\begin{figure}[t]
\begin{center}
\includegraphics[width=\linewidth]{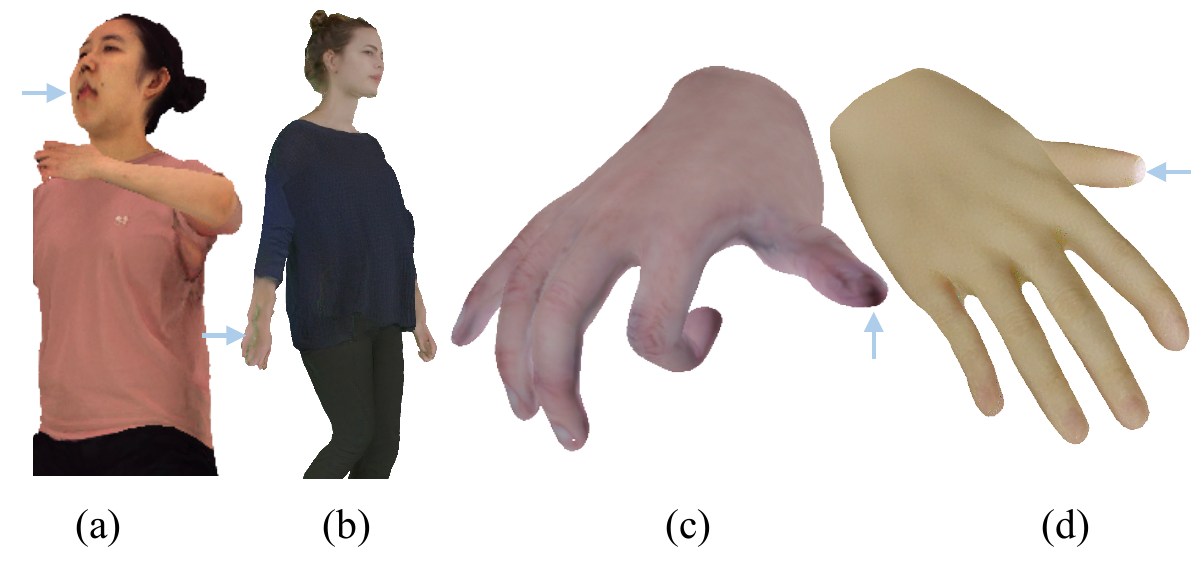}
\caption{Demonstration of limitations.
}
\label{fig:limit}
\end{center}
\vspace{-0.3cm}
\end{figure}

\vspace{-0.2cm}
\paragraph{Disentanglement of albedo and illumination.} 
Our framework generates human texture with mixed albedo and illumination, leading to errors in texture reconstruction. As shown in Fig.~\ref{fig:limit}(c), some black patterns appear on the top of fingers, which is caused by shadows in the training data. That is, due to a lack of awareness of lighting, the SDS guidance tends to generate shadow-like patterns in unseen regions. To address this issue, we plan to introduce illumination-aware designs (\eg, \cite{bib:HandAvatar}) to the HaveFun framework. 

\vspace{-0.2cm}
\paragraph{Full body integration with part-wise few-shot data.} 
This paper streamlines the data collection process and proves that few-shot unconstrained images are cheaper data sources for human avatar creation. In addition, we demonstrate that such a cheap data source is effective for the human body and hand. Nevertheless, we have not used the HaveFun framework for expressive portrait reconstruction. On one hand, because of the aforementioned limitations on facial expression, the HaveFun framework has difficulty in precise expression modeling. In addition, enhancing the accuracy of expression control is far from sufficient for modeling the portrait. For example, because of the lack of inner-mouth regions in the few-shot training data, the avatar is unable to perform a behavior with an open mouth (see Fig.~\ref{fig:limit}(a)). Therefore, we will explore a few-shot unconstrained data setup for portrait reconstruction. Finally, the portrait, body, and hand can be reconstructed from part-wise few-shot data and integrated into a full representation for an expressive human avatar.

\vspace{-0.2cm}
\paragraph{Errors caused by data pre-processing.} 
As illustrated in Fig.~\ref{fig:limit}, inaccurate image matting results in the introduction of background color to the human texture (Fig.~\ref{fig:limit}(b)). Additionally, artifacts such as the top of thumb could come from an inaccurate MANO/SMPLX fitting (Fig.~\ref{fig:limit}(d)).






\end{document}